%% file: arxiv.tex
\documentclass[letterpaper]{article} 
\usepackage{aaai2026}  
\usepackage{times}  
\usepackage{helvet}  
\usepackage{courier}  
\usepackage[hyphens]{url}  
\usepackage{graphicx} 
\usepackage{natbib}  
\usepackage{caption} 
\usepackage{algorithm}
\usepackage{algorithmic}

\usepackage{booktabs}       
\usepackage{amsfonts}       
\usepackage{nicefrac}       
\usepackage{microtype}      
\usepackage{xcolor}         

\usepackage{xspace} 
\usepackage{amsmath}
\usepackage{amssymb}
\usepackage{multirow}
\usepackage{multicol}
\usepackage{graphicx}

\usepackage{pifont}

\usepackage{float}  
\usepackage{tikz}
\usepackage{pgfplots}
\pgfplotsset{compat=1.18}
\usepackage{subcaption}
\usepackage[capitalise]{cleveref}

\usepackage[english]{babel}
\newtheorem{theorem}{Theorem}

\newcommand{\methodname}{ScoreAug\xspace}

\usepackage{newfloat}
\usepackage{listings}
\DeclareCaptionStyle{ruled}{labelfont=normalfont,labelsep=colon,strut=off} 
\floatstyle{ruled}
\newfloat{listing}{tb}{lst}{}
\floatname{listing}{Listing}
\nocopyright

\title{Score Augmentation for Diffusion Models}
\author{
    Liang Hou\textsuperscript{\rm 1}, Yuan Gao\textsuperscript{\rm 1}, Boyuan Jiang\textsuperscript{\rm 1}, Xin Tao\textsuperscript{\rm 1},\\Qi Yan\textsuperscript{\rm 2}, Renjie Liao\textsuperscript{\rm 2}, Pengfei Wan\textsuperscript{\rm 1}, Di Zhang\textsuperscript{\rm 1}, Kun Gai\textsuperscript{\rm 1}
}
\affiliations{
    \textsuperscript{\rm 1}Kling Team, Kuaishou Technology\quad \textsuperscript{\rm 2}University of British Columbia\\

    \{lianghou96, jiangsutx\}@gmail.com
}

\usepackage{bibentry}

\begin{document}

\maketitle

\begin{abstract}
Diffusion models have achieved remarkable success in generative modeling. However, this study confirms the existence of overfitting in diffusion model training, particularly in data-limited regimes. To address this challenge, we propose Score Augmentation (\methodname), a novel data augmentation framework specifically designed for diffusion models. Unlike conventional augmentation approaches that operate on clean data, \methodname applies transformations to noisy data, aligning with the inherent denoising mechanism of diffusion. Crucially, \methodname further requires the denoiser to predict the augmentation of the original target. This design establishes an equivariant learning objective, enabling the denoiser to learn scores across varied denoising spaces, thereby realizing what we term score augmentation. We also theoretically analyze the relationship between scores in different spaces under general transformations. In experiments, we extensively validate \methodname on multiple benchmarks including CIFAR-10, FFHQ, AFHQv2, and ImageNet, with results demonstrating significant performance improvements over baselines. Notably, \methodname effectively mitigates overfitting across diverse scenarios, such as varying data scales and model capacities, while exhibiting stable convergence properties. Another advantage of \methodname over standard data augmentation lies in its ability to circumvent data leakage issues under certain conditions. Furthermore, we show that \methodname can be synergistically combined with traditional data augmentation techniques to achieve additional performance gains.
\end{abstract}


\section{Introduction}

Diffusion models~\cite{ddpm,ncsn} have emerged as a powerful generative modeling approach, achieving remarkable success in tasks such as image generation~\cite{beatgan,ldm}.
Compared to GANs~\cite{gan,adcgan}, which rely on intricate adversarial training dynamics, diffusion models adopt a simpler learning objective rooted in iterative denoising. This simplicity endows diffusion models with stable optimization behavior and robust fitting capabilities. Recent studies reveal that diffusion models trained on identical datasets converge to nearly indistinguishable score functions, resulting in homogenized generation outputs~\cite{diffusion_memorization}. While this observation underscores their strong capacity to model data distributions, it also hints a critical issue: the strong fitting capacity of diffusion models may inadvertently amplify overfitting risks, particularly in data-limited regimes.

Despite widespread adoption, the overfitting issue in diffusion models remains underexplored.
Our empirical analysis confirms that diffusion models exhibit pronounced overfitting, especially when training data is limited or model capacity is excessive.
While regularization like dropout~\cite{dropout} or weight decay~\cite{weight_decay,adamw} offer partial mitigation, they often compromise the capability.
Data augmentation~\cite{autoaugment}, a cornerstone of enhancing the generalization of discriminative models, also offers a promising avenue to improve generative models~\cite{distaug,diffaug,augselfgan}.
However, conventional augmentation approaches designed for clean data do not account for the unique training dynamics of diffusion models, in which the denoiser operates on noisy data~\cite{edm,ddpm,ncsn}.
Moreover, heuristic data augmentation risks introducing distribution shifts, which may necessitate additional conditioning mechanisms to disentangle augmented data distributions.
These limitations underscore the urgent need for a principled framework tailored to the inherent mechanisms of diffusion models.

To address these issues, we propose Score Augmentation (\methodname), a novel augmentation framework that operates synergistically with the diffusion process.
Unlike conventional data augmentation methods that exclusively target clean data, \methodname applies transformations to noisy data.
Crucially, \methodname reformulates the denoising objective to predict augmented targets, thereby establishing an equivariant learning signal.
Specifically, distinct augmented data correspond to uniquely transformed noise patterns, effectively mitigating potential data leakage risks arising from noise invariance.
Our theoretical analysis reveals that this mechanism enables the denoiser to learn score across various transformation spaces with specific correspondence with the original score, hence the name score augmentation.
Extensive experiments on CIFAR-10, FFHQ, AFHQv2, and ImageNet across UNet~\cite{edm} and DiT~\cite{dit,sit} show that \methodname achieves significant performance and anti-overfitting improvements over baselines.
Notably, \methodname exhibits synergistic compatibility with conventional augmentation approaches, delivering cumulative performance gains when combined.

\section{Related Work}

\paragraph{Diffusion Models}
Diffusion models have recently emerged as a powerful framework for generative modeling.
The foundational work of DDPM~\cite{ddpm,noneq} established discrete-time diffusion processes with variational training, while DDIM~\cite{ddim} introduced deterministic and fast sampling through non-Markovian trajectories.
On the other hand, NCSN~\cite{ncsn} proposed noise conditional score matching to learn the Stein score~\cite{score} of perturbed data distribution at multiple noise levels.
Annealed Langevin dynamics~\cite{roberts1996exponential} is then used to sample from the noise~\cite{song2020improved}.
A unifying perspective emerged through stochastic differential equations (SDEs)~\cite{sde}, which generalized DDPM and NCSN to continuous-time dynamics.
The SDE framework categorizes diffusion processes into variance preserving (VP), variance exploding (VE) formulations, and sub-VP.
Subsequent works such as flow matching~\cite{flow_matching} and rectified flow~\cite{rectified_flow} can be seen as the sub-VP form.
EDM~\cite{edm} further unified the formulations of VP, VE and sub-VP under a single training framework with disentangled sampling parameters, later refined in EDM2~\cite{edm2} for enhanced training dynamics. 
On the variational perspective, VDM~\cite{vdm} and its improved variant VDM++~\cite{vdmpp} established theoretical connections to maximum likelihood estimation.
Recent breakthroughs in DiTs~\cite{dit,sit} demonstrate how transformer architectures can replace traditional U-Net backbones~\cite{ldm}, achieving state-of-the-art results~\cite{sd3}.
In terms of generalizability of diffusion models, recent works~\cite{diffusion_replication,diffusion_data} reveal that diffusion models tend to memorize training data when model capacity exceeds dataset size, raising concerns about replication risks.
Theoretical analyses~\cite{diffusion_generalization} establish polynomial relationships between generalization error bounds and sample size and model capacity. Complementary empirical study~\cite{on_diffusion_generalization} quantify memorization through mutual information, revealing that empirically optimal models often exhibit poor generalization.

\paragraph{Data Augmentation for Generative Models}
The concept of data augmentation in generative modeling was systematically explored by DistAug~\cite{distaug}, later widely adopted in GAN frameworks during their prominence. DiffAugment~\cite{diffaug} introduced differentiable augmentations specifically optimized for GAN training, while AugSelf-GAN~\cite{augselfgan} enhanced data efficiency through integrated self-supervised tasks. StyleGAN-ADA~\cite{ada} systematically analyzed the effects of augmentation in limited data regimes, with subsequent improvements in the probability of adaptive augmentation via APA~\cite{apa}. While these approaches primarily targeted GANs, recent diffusion models like EDM and EDM2 have successfully adapted conventional augmentation techniques through noise-conditional transformations.
Data augmentation naturally allows self-supervised tasks, such as equivariance constraints. In the field of diffusion models, AF-LDM~\cite{afldm} introduces shift-equivariance constraints to mitigate aliasing, while EquiVDM~\cite{equivdm} explores temporal equivariance in video diffusion models. However, none of these studies focus on addressing the overfitting issue in diffusion models.
From the theoretical aspect, \cite{score_cov} analyzes the change of variables for score in different spaces under invertible transformations, while our proposed \methodname also adopts and analyzes the noninvertible transformations.

\section{Preliminaries}

\input{figures/dm_da_sa}

Diffusion models are a family of generative models that consist of a forward noising and a backward denoising process.
The forward process of diffusion models typically corresponds to the following forward stochastic differential equation (SDE)~\cite{sde}:
\begin{equation*}
    \mathrm{d}\mathbf{x} = f(\mathbf{x},t)\mathrm{d}t + g(t)\mathrm{d}\mathbf{w},
\end{equation*}
where the drift coefficient has the form of $f(\mathbf{x},t)=f(t)\mathbf{x}$ with $f(t):\mathbb{R}\to\mathbb{R}$, and $g(t):\mathbb{R}\to\mathbb{R}$ is the diffusion coefficient. Here, $t\in[0,1]$ is the diffusion time, $\mathbf{x}\in\mathbb{R}^d$ is the data in the $d$-dimensional ambient space, and $\mathbf{w}\in\mathbb{R}^d$ is the standard Wiener process. The perturbation kernels of the SDE have the following form~\cite{edm}:
\begin{equation}\label{eq:edm_kernel}
    p(\mathbf{x}_t|\mathbf{x}_0) = \mathcal{N}(\mathbf{x}_t;s(t)\mathbf{x}_0,s(t)^2\sigma(t)^2\mathbf{I}),
\end{equation}
where $\mathcal{N}(\mathbf{x};\boldsymbol{\mu},\mathbf{\Sigma})$ is the probability density function of Gaussian distribution with mean $\boldsymbol{\mu}\in\mathbb{R}^d$ and covariance $\mathbf{\Sigma}\in\mathbb{R}^{d\times d}$ evaluated at data point $\mathbf{x}$. The coefficients are:
\begin{equation*}
    s(t)=\exp\left(\int_0^t f(\xi)\mathrm{d}\xi\right) \ \text{and}\ \sigma(t)=\sqrt{\int_0^t\frac{g(\xi)^2}{s(\xi)^2}\mathrm{d}\xi}.
\end{equation*}
According to the different choices of the monotonically decreasing schedule $s(t)$ and the monotonically increasing schedule $\sigma(t)$, diffusion models can be divided into three formulations: 1) variance exploding (VE) $s(t)=1, \sigma(t)=\sqrt{t}$ such that $s(t)^2+s(t)^2\sigma(t)^2>1, \forall t\in(0,1]$; 2) variance preserving (VP), $s(t)^2+s(t)^2\sigma(t)^2=1, \forall t\in[0,1]$; 3) sub-VP, $s(t)^2+s(t)^2\sigma(t)^2\leq1, \forall t\in[0,1]$.
The forward SDE corresponds to the probability flow ordinary differential equation (PF-ODE)~\cite{sde,edm} that can recover the data distribution $p_{\text{data}}$ from the tractable prior $\mathcal{N}(\mathbf{0},s(1)^2\sigma(1)^2\mathbf{I})$:
\begin{equation*}
    \mathrm{d}\mathbf{x} = \left[\frac{\dot s(t)}{s(t)}\mathbf{x}-s(t)^2\dot\sigma(t)\sigma(t)\nabla_{\mathbf{x}}\log p\left(\frac{\mathbf{x}}{s(t)};\sigma(t)\right)\right]\mathrm{d}t,
\end{equation*}
where we have $p(\mathbf{x};\sigma) \triangleq p_{\text{data}} * \mathcal{N}(\mathbf{0},\sigma(t)^2\mathbf{I})$ with $*$ the convolution operator, so that the marginal distribution of perturbed data at time $t$ is $p(\mathbf{x}_t)=s(t)^{-d}p(\mathbf{x}_t/s(t);\sigma(t))$.
EDM~\cite{edm} leverages a denoiser function $D(\cdot;\sigma):\mathbb{R}^d\to\mathbb{R}^d$ for noise level $\sigma(t)$\footnote{We follow EDM to omit $t$ for simplicity when there is no ambiguity in the context.} with the optimization objective:
\begin{equation}\label{eq:edm_loss}
    \mathcal{L}_{\mathrm{edm}}(D;\sigma) = \mathbb{E}_{\mathbf{d}\sim p_{\text{data}},\mathbf{n}\sim\mathcal{N}(\mathbf{0},\sigma^2\mathbf{I})}\|D(\mathbf{d}+\mathbf{n};\sigma)-\mathbf{d}\|_2^2,
\end{equation}
where the denoiser is typically constructed as $D_{\boldsymbol{\theta}}(\mathbf{x};\sigma)=c_{\text{skip}}(\sigma)\mathbf{x}+c_{\text{out}}(\sigma)F_{\boldsymbol{\theta}}(c_{\text{in}}(\sigma)\mathbf{x};c_{\text{noise}}(\sigma))$ with predefined scaling functions $c_{\text{skip}}, c_{\text{out}}, c_{\text{in}}, c_{\text{noise}}:\mathbb{R}_{\geq 0}\to\mathbb{R}$ and a neural network $F_{\boldsymbol{\theta}}:\mathbb{R}^d\times\mathbb{R}\to\mathbb{R}^d$ with trainable parameters $\boldsymbol{\theta}\in\Theta$.
The Stein score (the gradient of log density of the perturbed data distribution $p(\mathbf{x};\sigma)$) can be obtained from the optimal denoiser:
\begin{equation}\label{eq:edm_score}
    \nabla_{\mathbf{x}}\log p(\mathbf{x};\sigma) = \frac{D(\mathbf{x};\sigma)-\mathbf{x}}{\sigma^2}.
\end{equation}

\section{Method}
\label{sec:method}

\input{tables/main_results}

\subsection{Linear Transformations}

According to the perturbation kernel (\cref{eq:edm_kernel}) in diffusion models, the forward process can be expressed as:
\begin{equation}\label{eq:edm_forward}
    \mathbf{x}_t = s(t)\mathbf{x}_0 + s(t)\sigma(t)\boldsymbol{\epsilon},
\end{equation}
where $\boldsymbol{\epsilon}\sim \mathcal{N}(\mathbf{0},\mathbf{I})$ is the normal multivariate Gaussian noise.
Let augmentation $T(\cdot,\boldsymbol{\omega}):\mathbb{R}^d\to\mathbb{R}^d$, with parameter $\boldsymbol{\omega}\in\boldsymbol{\Omega}$ that determine the augmentation transformation, be a linear transformation (see general case later) that has the form of $T(\mathbf{x},\boldsymbol{\omega})=\mathbf{T}_{\boldsymbol{\omega}}\mathbf{x}$ for the corresponding transformation matrix $\mathbf{T}_{\boldsymbol{\omega}}\in\mathbb{R}^{d\times d}$.\footnote{For notational simplicity, we also omit $\boldsymbol{\omega}$ in $T(\cdot;\boldsymbol{\omega})$ and $\mathbf{T}_{\boldsymbol{\omega}}$ when the context is unambiguous.}
Each augmentation defines an augmented space, and the forward process in the augmented space is:
\begin{equation}\label{eq:scoreaug_forward}
    \mathbf{y}_t\triangleq T(\mathbf{x}_t;\boldsymbol{\omega}) = s(t)\mathbf{T}\mathbf{x}_0+s(t)\sigma(t)\mathbf{T}\boldsymbol{\epsilon},
\end{equation}
According to this forward process, we can obtain the perturbation kernel in the augmented space:
\begin{equation*}
    p(\mathbf{y}_t|\mathbf{y}_0) = \mathcal{N}(\mathbf{y}_t;s(t)\mathbf{y}_0,s(t)^2\sigma(t)^2\mathbf{T}\mathbf{T}^\top).
\end{equation*}
For finite data samples $\{\mathbf{x}_1,\dots,\mathbf{x}_N\}$, where $N\in\mathbb{Z}^+$ is the number of observed training data, the empirical data distribution can be constructed as $p_\mathrm{data}(\mathbf{x})=1/N\sum_{i=1}^N\delta(\mathbf{x}-\mathbf{x}_i)$ with standard deviation $\sigma_\text{data}\in\mathbb{R}^+$, where $\delta(\cdot)$ is the Dirac delta function.
Let the corresponding transformed data samples be $\{\mathbf{y}_1,\dots,\mathbf{y}_N\}$ with the transformed data density $\hat{p}_\mathrm{data}(\mathbf{y})=1/N\sum_{i=1}^N\delta(\mathbf{y}-\mathbf{y}_i)$ under a given augmentation.
And the distribution of the transformed data at time $t$ is $p_t(\mathbf{y}) = \iint_{\mathbb{R}^d\times\mathbb{R}^d}\delta(\mathbf{y}-T(\mathbf{x}_t;\boldsymbol{\omega}))p(\mathbf{x}_t|\mathbf{x}_0)p(\mathbf{x}_0)\mathrm{d}\mathbf{x}_t\mathrm{d}\mathbf{x}_0 = \int_{\mathbb{R}^d} p(\mathbf{y}_t|\mathbf{y}_0) p(\mathbf{y}_0) \mathrm{d}\mathbf{y}_0$.
We can then define the distribution of the transformed data at noise level $\sigma$ as:
\begin{equation*}
    p(\mathbf{y};\sigma) = \hat{p}_{\text{data}} * \mathcal{N}(\mathbf{0},\sigma(t)^2\mathbf{T}\mathbf{T}^\top) =s(t)^d p_{t^{-1}(\sigma)}(s(t)\mathbf{y}).
\end{equation*}
By analogy with the relationship between the original forward process (\cref{eq:edm_forward}) and the denoiser loss function (\cref{eq:edm_loss}) -- namely, that the original denoiser $D$ takes $\mathbf{\hat{x}}_t$ as input to predict $\mathbf{x}_0$ -- it is reasonable to assume that, given the new forward process (\cref{eq:scoreaug_forward}), the new denoiser is expected to take $\mathbf{\hat{y}}_t$ as input to predict $\mathbf{y}_0$.
This observation motivates us to design the loss function for each noise level and augmentation of the augmented denoiser as follows:
\begin{equation}\label{eq:scoreaug_loss}
    \mathcal{L}(D;\sigma,\boldsymbol{\omega}) = \mathbb{E}_{\mathbf{d},\mathbf{n}}\|D(T(\mathbf{d}+\mathbf{n};\boldsymbol{\omega});\sigma,\boldsymbol{\omega}) - T(\mathbf{d};\boldsymbol{\omega})\|_2^2.
\end{equation}
And the total loss function is calculated expected on all augmentations and all noise levels: $\mathcal{L}(D)=\mathbb{E}_{\boldsymbol{\omega}\sim p_\Omega}\mathbb{E}_{\sigma\sim p_\sigma}\lambda(\sigma)\mathcal{L}(D;\sigma,\boldsymbol{\omega})$,
where $p_\sigma$ is the prior of noise level (we follow EDM to set $\ln(\sigma)\sim \mathcal{N}(-1.2,1.2^2)$) and $p_\Omega$ is the prior of augmentation parameters (see below), and $\lambda(\sigma)=(\sigma^2+\sigma_\text{data}^2)/(\sigma\cdot\sigma_\text{data})^2$ is the loss weighting that also follows EDM.
Under the assumption of infinite model capacity, we can prove (see Appendix) that the ideal augmented denoiser has the form of:
\begin{equation*}
    D(\mathbf{y};\sigma,\boldsymbol{\omega})=\frac{\sum_{i=1}^Y\mathcal{N}(\mathbf{y};\mathbf{y}_i,\sigma^2\mathbf{T}\mathbf{T}^\top)\mathbf{y}_i}{\sum_{i=1}^Y\mathcal{N}(\mathbf{y};\mathbf{y}_i,\sigma^2\mathbf{T}\mathbf{T}^\top)}.
\end{equation*}
Thus, the score with respect to augmented data $\mathbf{y}$ at noise level $\sigma$ can be obtained from the ideal augmented denoiser:
\begin{equation}\label{eq:scoreaug_score}
    \nabla_\mathbf{y}\log p(\mathbf{y};\sigma) = (\mathbf{T}\mathbf{T}^\top)^\dagger\left(D(\mathbf{y};\sigma,\boldsymbol{\omega})-\mathbf{y}\right)/\sigma^2,
\end{equation}
where $(\mathbf{T}\mathbf{T}^\top)^\dagger$ is the Moore-Penrose inverse of $\mathbf{T}\mathbf{T}^\top$. If $\mathbf{TT}^{\top}$ is a singular matrix, the corresponding Gaussian distribution is a degenerate distribution. The above formula also holds when the gradient $\nabla_{\mathbf{y}} \log p(\mathbf{y};\sigma)$ is defined in the image space $\mathrm{Im}(\mathbf{T})$ of the matrix $\mathbf{T}$.
From this perspective, it becomes evident that \methodname essentially requires the optimal denoiser to be equivariant with respect to the employed linear transformation, as written as:
\begin{equation*}
    D(\mathbf{T}\mathbf{x};\sigma)=\mathbf{T}\mathbf{x}+\sigma^2\mathbf{T}\mathbf{T}^\top\nabla_{\mathbf{T}\mathbf{x}}\log p(\mathbf{T}\mathbf{x};\sigma)=\mathbf{T}D(\mathbf{x};\sigma).
\end{equation*}
For $\mathbf{y}=T(\mathbf{x})$, where $T$ is linear and invertible, we have $\nabla_{\mathbf{y}} \log p(\mathbf{y};\sigma)=\mathbf{T}^{-\top}\nabla_{\mathbf{x}}\log p(\mathbf{x};\sigma)$ that reveals the correspondence between scores in transformed spaces (see~\cref{thm:score_transformation} for general case). When combined with~\cref{eq:edm_score,eq:scoreaug_score}, this demonstrates that the new denoiser learns scores in different spaces, indicating that \methodname performs score augmentation rather than data augmentation.

\paragraph{Augmentation and Condition}
We borrow the practices of data-augmented GANs~\cite{diffaug,augselfgan,ada} to adopt linear transformations (brightness, translation, cutout, and rotation) as the data transformation (see~\cref{eq:nonlinear_loss} for nonlinear case). We empirically find that any independent augmentation can improve the performance of the baseline, and the combination is significantly better (see~\cref{tab:different_augmentations}).

Let $H\in\mathbb{Z}^+$ and $W\in\mathbb{Z}^+$ be the height and width of the image, respectively. The transformation matrix corresponding to different augmentations can be defined as follows.
\begin{itemize}
    \item \textbf{Brightness} scales the images by $\omega_b\in[1/B, B]$, such that $T^{\omega_b}_{ij}=\{\omega_b \text{ if } i=j,\ 0 \text{ otherwise}\}.$
    \item \textbf{Translation} shifts images by $\Delta_i\in\{1,\cdots,\lfloor R_t W\rfloor\}$ vertical and $\Delta_j\in\{1,\cdots,\lfloor R_t H\rfloor\}$ horizontal pixels, with $\boldsymbol{\omega}_t=(\Delta_i, \Delta_j)$, such that $T^{\boldsymbol{\omega}_t}_{ij}=\{
        1 \text{ if } i=j+\Delta i+H\cdot \Delta j,\ 0 \text{ otherwise}\}.$
    \item \textbf{Cutout} zeros a rectangular region centered at point of $(c_x,c_y)$ with size of $(h,w)$ that $h\in\{1,\cdots,R_c H\},w\in\{1,\cdots,R_c W\}$, where $\boldsymbol{\omega}_c=(c_x,c_y,h,w)$, such that $
    T^{\boldsymbol{\omega}_c}_{ij}=\{
        1 \text{ if } \left| \frac{i}{W} - c_y \right| > \frac{h}{2} \text{ or } \left| (i \bmod W) - c_x \right| > \frac{w}{2},\ 0 \text{ otherwise}\}.$
    \item \textbf{Rotation} rotates images by $90^\circ \times \omega_r$, where $\omega_r\in\{0,1,2,3\}$, such that $T^{\omega_r}_{ij}=\{1 \text{ if } i=(H-1-h)+wW, j=h+wH,\ 0 \text{ otherwise}\}.$
\end{itemize}

Note that translation and cropping are zero-padded instead of masked. The difference is that we calculate the loss of the padded area, while the mask does not.
The augmentation parameters are randomly sampled from predetermined ranges that include identity mapping to ensure learning from the original data.
For the condition input (if any) to the denoiser, \methodname add a linear layer to directly accept the condition vector $\boldsymbol{\omega}$, and then add it together with the timestep embedding.
For cutout, the center point coordinate $(c_x, c_y)$ is removed and only the cutout size $\boldsymbol{\omega}_c=(h,w)$ is kept.
When sampling, we can set the condition (if any) to an appropriate value (e.g., zeros) to generate an untransformed image and prevent augmentation-leaking for aggressive augmentations (see~\cref{tab:condition}).

\input{tables/dit}

\input{tables/different_augmentations}

\subsection{Extension to Nonlinear Transformations}\label{sec:nonlinear}
While the preceding discussion has focused on linear transformation-based data augmentation, the proposed method can be naturally extended to nonlinear scenarios. Notably, the loss function (\cref{eq:scoreaug_loss}) imposes no inherent restrictions on the linearity of the transformation $T$. In our experiments, we therefore explore directly adopting all data augmentation techniques from EDM as candidate implementations of $T$. Additionally, we introduce an alternative loss function formulation where transformations are separately applied to the data and noise inputs, with their combined output fed into the denoiser to predict the equivalently transformed data (\cref{eq:nonlinear_loss}). These two loss functions exhibit equivalence under linear transformation, yet demonstrate divergence in more generalized settings, with comparative performance provided in~\cref{tab:main_results}.
\begin{align}\label{eq:nonlinear_loss}
    \mathcal{L}(D;\sigma,\boldsymbol{\omega}) = \mathbb{E}_{\mathbf{d},\mathbf{n}}\|D(T(\mathbf{d})+T(\mathbf{n});\sigma) - T(\mathbf{d})\|_2^2.
\end{align}

Back to the augmentation form of~\cref{eq:scoreaug_loss}, the following theorem establishes the correspondence between score functions in different spaces under general transformations.
\begin{theorem}[Transformation of Score Functions]\label{thm:score_transformation}
Let $p(\mathbf{x};\sigma)$ be the probability density function (PDF) of $\mathbf{x} \in \mathbb{R}^n$. Let $\mathbf{y}=T(\mathbf{x})$ where $T:\mathbb{R}^n \to \mathbb{R}^m (m\leq n)$ is a differentiable map with Jacobian $\mathbf{J}_T(\mathbf{x})$ and $p(\mathbf{y};\sigma)$ is the PDF of $\mathbf{y}$. Assuming sufficient smoothness and positivity for $p(\mathbf{x};\sigma)$, $T(\mathbf{x})$, and $p(\mathbf{y};\sigma)$ such that all terms below are well-defined, if $\mathbf{J}_T(\mathbf{x})$ has full row rank $m$ for $\mathbf{x}$ in the support of $p(\mathbf{x}|\mathbf{y};\sigma)$, then:
\begin{align*}
 &\nabla_{\mathbf{y}} \log p(\mathbf{y};\sigma) = \\ & \mathbb{E}_{p(\mathbf{x}|\mathbf{y};\sigma)} \left[ \mathbf{J}_T(\mathbf{x})^{\dagger}\left( \nabla_{\mathbf{x}} \log p(\mathbf{x};\sigma) -\frac{1}{2} \nabla_{\mathbf{x}} \log \mathcal{J}(\mathbf{x}) \right) \right],
\end{align*}
where $\mathbf{J}_T(\mathbf{x})^{\dagger} \triangleq (\mathbf{J}_T(\mathbf{x})\mathbf{J}_T(\mathbf{x})^\top)^{-1}\mathbf{J}_T(\mathbf{x})$ and $\mathcal{J}(\mathbf{x})=\det(\mathbf{J}_T(\mathbf{x})\mathbf{J}_T(\mathbf{x})^\top)$.
The proof is deferred to Appendix.
\paragraph{Diffeomorphism} If $m=n$ and $T$ is a (global) diffeomorphism with $\mathbf{x}= T^{-1} (\mathbf{y})$:
\begin{align*}
& \nabla_{\mathbf{y}} \log p (\mathbf{y};\sigma) \\ &= \mathbf{J}_T(\mathbf{x})^{-\top}\left( \nabla_{\mathbf{x}} \log p(\mathbf{x};\sigma) - \nabla_{\mathbf{x}} \log |\det(\mathbf{J}_T(\mathbf{x}))| \right)
\end{align*}
\paragraph{Linear Surjection} If $T(\mathbf{x})=\mathbf{Tx}$, where $\mathbf{T}$ is a constant $m\times n$ matrix with full row rank $m$:
\begin{equation*}
\nabla_{\mathbf{y}} \log p(\mathbf{y};\sigma) = (\mathbf{T}\mathbf{T}^\top)^{-1}\mathbf{T} \cdot \mathbb{E}_{p(\mathbf{x}|\mathbf{y};\sigma)} \left[\nabla_{\mathbf{x}} \log p(\mathbf{x};\sigma)\right]
\end{equation*}
\end{theorem}

\section{Experiments}

\subsection{Experimental Setup}
\label{sec:exp_setup}

We implement the proposed \methodname based on the official EDM code\footnote{\url{https://github.com/NVlabs/edm}} due to its generality.
We follow the settings of EDM, including the network and preconditioning, training, sampling, and parameters (see Table 1 in the EDM paper~\cite{edm}), to conduct experiments for a fair comparison with the baselines.
The datasets are unconditional CIFAR-10 and conditional CIFAR-10~\cite{cifar}, FFHQ~\cite{ffhq}, and AFHQv2~\cite{afhq}.
On each dataset, we experiment with both variance preserving (VP)~\cite{iddpm,edm} and variance exploding (VE)~\cite{ncsn,edm} formulations.
Fréchet Inception Distance (FID)~\cite{fid} is used as the main evaluation metric.
All models are trained for 200,000 iterations with batch size of 512.
All results are calculated from the model evaluation at the last scheduled checkpoint for fair comparisons with 50,000 generated images unless otherwise specified.
In the experiments, we use the code officially provided by EDM and re-run its results as a baseline for an absolutely fair comparison.
For the specific data augmentation types finally used by each method, please refer to Appendix.

\subsection{Main Results}
\label{sec:main_results}

\cref{tab:main_results} presents comparative results between our method, the baseline, and competing methods. The baseline, denoted as EDM w/o NLA (without non-leaking augmentation), exhibits clear signs of overfitting, as evidenced by performance gains when increasing dropout or incorporating weight decay. In contrast, integrating \methodname significantly outperforms these basic overfitting mitigation strategies, underscoring the efficacy of our approach.
Notably, EDM w/ NLA, which employs sophisticated non-linear augmentations, achieves superior results. Remarkably, \methodname's non-linear extension can seamlessly generalize to these augmentations. Both variants (\cref{eq:scoreaug_loss,eq:nonlinear_loss}) attain improved FID scores across most scenarios, demonstrating their expansibility.
Furthermore, the linear variant of \methodname can be applied synergistically to EDM w/ NLA, yielding additional performance improvements. This flexibility highlights the broad applicability of our method, even when integrated with state-of-the-art augmentation frameworks.
\cref{fig:afhq} and (Fig.~6 in Appendix) in show the images generated by EDM and \methodname trained on AFHQ-v2 and FFHQ, respectively.

\input{tables/condition}

\paragraph{Diffusion Transformer Architecture} We also conducted experiments on ImageNet-256~\cite{imagenet} using the diffusion transformer architecture~\cite{dit}, adopting the state-of-the-art SiT~\cite{sit} training code\footnote{\url{https://github.com/willisma/SiT}} with an XL-scale model configuration. Evaluation metrics include FID~\cite{fid}, sFID~\cite{sfid}, IS~\cite{is}, as well as precision and recall~\cite{precision_recall}. As evidenced in~\cref{tab:dit}, \methodname demonstrates superior performance over SiT-XL in most metrics at 400K and 1M training steps, substantiating its effectiveness when applied to advanced model architectures and large complex datasets.

\subsection{Analysis Experiments}

\paragraph{Different Augmentations} 

Note that \methodname is compatible with multiple linear data augmentation types by randomly selecting one augmentation per training iteration. We investigated the impact of individual augmentations (brightness, translation, cutout, rotation) and their combinations on model performance. As shown in~\cref{tab:different_augmentations}, all single augmentations outperform the no-augmentation baseline, and combined usage achieves the best results, demonstrating synergistic effects. Based on the experimental results, we posit that incorporating more linear data augmentations can further improve performance, and we leave this exploration for future work.

\input{figures/afhq}

\paragraph{Importance of Conditioning}

In the method section introducing the \methodname method, we default to incorporating the augmentation parameter $\boldsymbol{\omega}$ as the condition for the model to distinguish between data augmentation types and intensities. However, this conditioning is not strictly necessary. As reported in~\cref{tab:condition}, \methodname remains functional without conditioning if the augmentations are non-invertible, such as translation and cutout. However, data augmentation without conditioning (DataAug w/o condition) causes data leakage in this case, leading to a significant drop in FID. Conversely, for invertible augmentations like uniform random rotation by $\{0^\circ, 90^\circ, 180^\circ, 270^\circ\}$~\cite{ada,ssganla}, unconditional \methodname fails to generate rotated images, resulting in augmentation-leaking issue. This failure arises because the random noise distribution is rotation-invariant, causing \methodname to treat rotated images as original training data, effectively reducing it to standard data augmentation. 

Based on these findings, we recommend adding conditions to support broader augmentation types, though this modifies the network architecture. For users aiming to finetune pre-trained models, unconditional \methodname remains viable if non-invertible augmentations are adopted. Additionally, conditional injection enables augmentation-controllable generation. For example, synthesizing images with specified rotation angles, as illustrated in~\cref{fig:rot_uncond_cifar10,fig:rot_cond_cifar10}.

\paragraph{Training Data} 
Overfitting typically stems from insufficient data and low data utilization efficiency of models. To validate this, we reduced the CIFAR-10 training data to $N=10,000$ and $N=20,000$ samples, comparing \methodname against the baseline. Results in~\cref{fig:a,fig:d} show that baseline performance degrades sharply with smaller datasets, while \methodname consistently outperforms it, demonstrating stronger data utilization efficiency and robustness against overfitting.

\paragraph{Model Size}
Another factor influencing model overfitting is model complexity -- larger models generally tend to overfit more easily. We compared \methodname and the baseline across varying model sizes by adjusting the number of base channels $C\in\mathbb{N}^+$, which is defaultly set as $C=128$ in EDM. We tested two additional configurations ($C=96$ and $C=160$) to decrease or increase the model size. As shown in~\cref{fig:b,fig:e}, the baseline EDM performance degrades with increasing model size, further indicating overfitting issues, while our \methodname improves significantly, highlighting its robustness against overfitting.

\input{figures/overfit}

\paragraph{Training Convergence}

To visually demonstrate the overfitting issue in diffusion models, we evaluated FID scores at each training checkpoint. In order to reduce the evaluation computational overhead, the FID scores are reduced to 10,000 generated images for calculation, which just matches the number of test images of CIFAR-10. As illustrated in~\cref{fig:c,fig:f}, the FID score of EDM initially decreases rapidly to a minimum and then gradually rises during training, indicating overfitting issues. In contrast, our \methodname maintains a consistent downward trend in FID scores, effectively mitigating overfitting throughout the training process. Although early stopping allows the baseline to achieve decent FID scores, it still underperforms \methodname at their best. Furthermore, early stopping necessitates continuous model evaluation, resulting in additional computational overhead. Overall, our method achieves more stable convergence properties.

\section{Conclusion}

In this paper, we identify the risk of overfitting in diffusion models and experimentally validate this phenomenon on data-limited regimes. To address this issue, we propose a novel data augmentation method tailored for diffusion models, termed Score Augmentation (\methodname). Unlike conventional augmentation approaches, \methodname applies transformations to noisy inputs (or data and noise, respectively). At this point, the diffusion model (denoiser) is trained to predict the data after the same transformation. This equivariant loss design ensures seamless integration of \methodname with the core objectives of diffusion models under linear transformations. Furthermore, we extend it to non-linear transformations and analyze the general relationship of score functions in different data spaces.
Experiments on benchmark image generation datasets (CIFAR-10, FFHQ, AFHQv2, and ImageNet) demonstrate that \methodname achieves significant improvements over the baseline. It exhibits robust resistance to overfitting under varied settings, including reduced training data and increased model complexity. In particular, \methodname can be combined with conventional overfitting mitigation strategies such as standard data augmentation to further improve performance.

\paragraph{Limitations and Broader Impacts}

The proposed method may exhibit slower convergence rates under limited data regimes, requiring longer training durations to reach the optimal checkpoint.
When combined with conventional overfitting mitigation techniques, the improvements may be less pronounced.
While the method works well in alleviating overfitting, it does not provide better convergence speed when the base model is still underfitting.
By improving generalization of diffusion models, the method could broaden real-world applications like medical imaging and creative design where training data is scarce.
However, mitigating memorization risks also helps prevent unintended data replication, supporting ethical AI development and copyright compliance in generated content.

\bibliography{main}


\newpage
\input{appendix}

\end{document}

%% file: figures/dm_da_sa.tex
\begin{figure*}[t]
\centering
\includegraphics[width=1.0\textwidth]{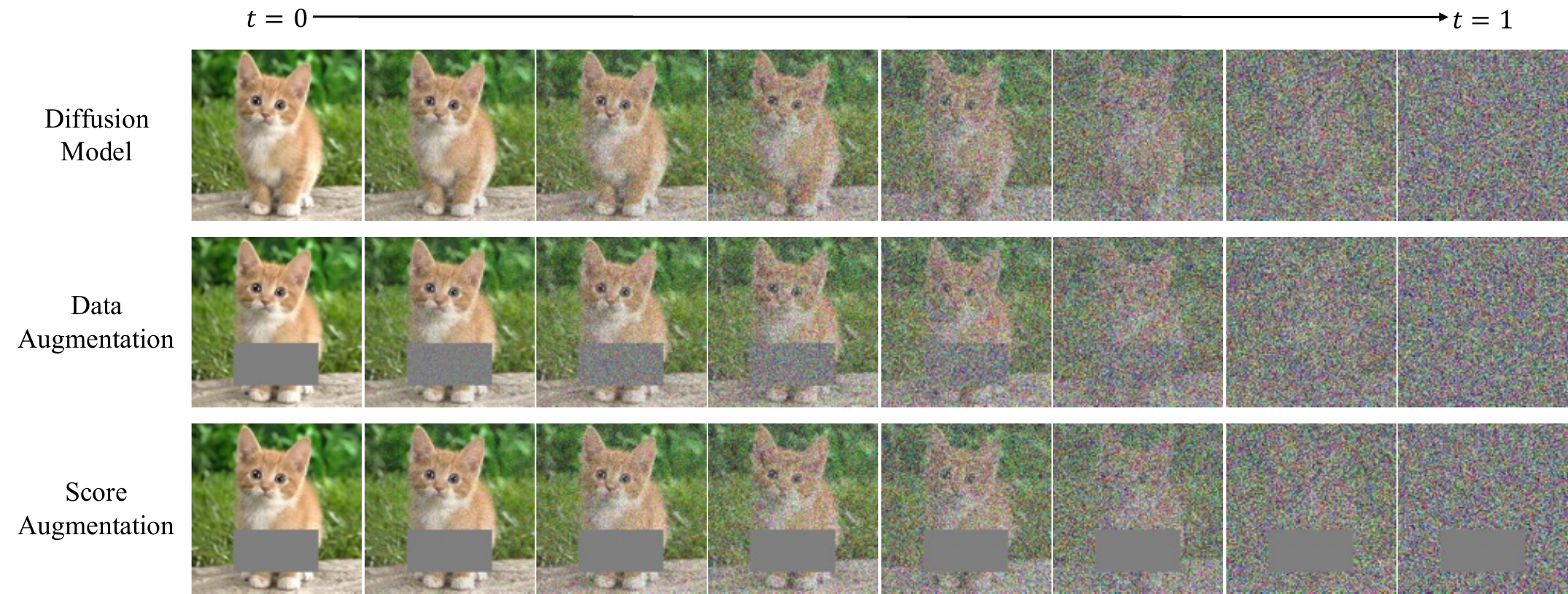}
\caption{Illustration of forward processes of diffusion models, data augmentation and \methodname.}
\label{fig:model}
\end{figure*}

%% file: tables/main_results.tex
\begin{table*}[t]
\caption{FID comparisons between \methodname and EDM on unconditional and conditional CIFAR-10, FFHQ, and AFHQv2. NLA means non-leaky augmentation on original data. VP and VE means variance-preserving and variance-exploding, respectively. NFE means the number of function evaluations. We reproduce the results of EDM using the official code for a fair comparison.}
  \label{tab:main_results}
  \centering
  \begin{tabular}{lcccccccc}
    \toprule
    \multirow{3}*{Method} & \multicolumn{4}{c}{CIFAR-10 $32\times32$} & \multicolumn{2}{c}{FFHQ $64\times64$} &  \multicolumn{2}{c}{AFHQv2 $64\times64$} \\
    \cmidrule(lr){2-5} \cmidrule(lr){6-7} \cmidrule(lr){8-9}
    & \multicolumn{2}{c}{Unconditional} & \multicolumn{2}{c}{Conditional} & \multicolumn{2}{c}{Unconditional} & \multicolumn{2}{c}{Unconditional} \\
    \cmidrule(lr){2-3} \cmidrule(lr){4-5} \cmidrule(lr){6-7} \cmidrule(lr){8-9}
    & VP & VE & VP & VE & VP & VE & VP & VE \\
    \midrule
    EDM w/o NLA
        & 4.05 & 4.10 & 4.03 & 4.32 
        & 5.26 & 4.98 & 5.69 & 5.58 \\
    + dropout$\times2$
        & 3.13 & 2.93 & 2.93 & 2.77
        & 4.87 & 4.63 & 4.60 & 4.54 \\
    + weight decay
        & 3.13 & 3.01 & 3.17 & 2.93 
        & 4.76 & 4.69 & 5.76 & 4.93 \\
    + \methodname
        & \textbf{2.35} & \textbf{2.24} & \textbf{2.11} & \textbf{2.25} 
        & \textbf{2.96} & \textbf{2.88} & \textbf{3.55} & \textbf{3.54} \\
    \midrule
    EDM w/ NLA
        & 2.07 & 2.10 & 1.93 & 1.92 
        & 2.76 & 2.80 & 2.65 & 2.68 \\
    + \methodname
        & \textbf{2.05} & 2.06 & \textbf{1.80} & 1.91 
        & 2.72 & \textbf{2.69} & \textbf{2.30} & \textbf{2.18} \\
    ScoreAug$\times$NLA~\cref{eq:scoreaug_loss} & \textbf{2.05} & \textbf{1.96} & 1.90 & \textbf{1.81} & \textbf{2.63} & 2.89 & 2.70 & 2.68 \\
    ScoreAug$\times$NLA~\cref{eq:nonlinear_loss} & 2.06 & 1.97 & 1.85 & 1.96 & 2.76 & 3.02 & 2.37 & 2.58 \\
    \midrule
    NFE & 35 & 35 & 35 & 35 & 79 & 79 & 79 & 79 \\
    \bottomrule
  \end{tabular}
\end{table*}

%% file: tables/dit.tex

\begin{table}[t]
  \caption{Quantitative comparisons between SiT and \methodname on ImageNet-256.}
  \label{tab:dit}
  \centering
  \begin{tabular}{lcccccc}
    \toprule
    Model & Steps & FID $\downarrow$ & sFID $\downarrow$ & IS $\uparrow$ & Precision $\uparrow$\\
    \midrule
    SiT-XL & 400K & 19.26 & 5.24 & 70.75 & 0.6223 \\
    + \methodname & 400K & \textbf{18.75} & \textbf{5.21} & \textbf{71.79} & \textbf{0.6249} \\
    \midrule
    SiT-XL & 1M & 13.21 & 5.39 & 94.64 & 0.6542 \\
    + \methodname & 1M & \textbf{12.70} & \textbf{5.36} & \textbf{96.37} & \textbf{0.6589} \\
    \bottomrule
    \end{tabular}
\end{table}

%% file: tables/different_augmentations.tex
\begin{table*}[t]
  \caption{Ablation study of ScoreAug under individual and combined augmentations. FID scores are reported for both unconditional and conditional CIFAR-10 settings across the variance preserving and variance exploding paradigms.}
  \label{tab:different_augmentations}
  \centering
  \begin{tabular}{lcccccccc}
    \toprule
    \multirow{2}{*}{Method} & \multicolumn{4}{c}{Augmentation} & \multicolumn{2}{c}{Uncond CIFAR} & \multicolumn{2}{c}{Cond CIFAR} \\
    \cmidrule(lr){2-5} \cmidrule(lr){6-7} \cmidrule(lr){8-9}
    & brightness & translation & cutout & rotation & VP & VE & VP & VE \\
    \midrule
    EDM & - & - & - & -
        & 4.05 & 4.10 & 4.03 & 4.32 \\
    + \methodname & \ding{51} & \ding{55} & \ding{55} & \ding{55}
        & 2.97 & 2.85 & 2.86 & 2.68 \\
    + \methodname & \ding{55} & \ding{51} & \ding{55} & \ding{55}
        & 2.68 & 2.86 & 2.40 & 2.62 \\
    + \methodname & \ding{55} & \ding{55} & \ding{51} & \ding{55}
        & 3.68 & 3.56 & 3.62 & 3.24 \\
    + \methodname & \ding{55} & \ding{55} & \ding{55} & \ding{51}
        & 2.43 & 2.69 & 2.13 & 2.59 \\
    + \methodname & \ding{51} & \ding{51} & \ding{51} & \ding{51}
        & \textbf{2.27} & \textbf{2.29} & \textbf{2.11} & \textbf{2.06} \\
    \bottomrule
    \end{tabular}
\end{table*}

%% file: tables/condition.tex
\begin{table*}[htbp]
  \caption{FID scores of ScoreAug without or with conditioning under different augmentation combination combinations on unconditional and conditional CIFAR-10 across VP and VE settings.}
  \label{tab:condition}
  \centering
  \begin{tabular}{lccccccc}
    \toprule
    \multirow{2}{*}{Method} & \multicolumn{3}{c}{Augmentation} 
        & \multicolumn{2}{c}{Uncond CIFAR} 
        & \multicolumn{2}{c}{Cond CIFAR} \\
    \cmidrule(lr){2-4} \cmidrule(lr){5-6} \cmidrule(lr){7-8}
    & translation & cutout & rotation & VP & VE & VP & VE \\
    \midrule
    EDM
        & - & - & - & 4.05 & 4.10 & 4.03 & 4.32 \\
    + DataAug w/o condition & \ding{51} & \ding{51} & \ding{55} & 11.02 & 10.87 & 10.31 & 10.36 \\
    + {\methodname} w/o condition       
        & \ding{51} & \ding{51} & \ding{55} & 2.53 & 2.63 & 2.41 & 2.37 \\
    + {\methodname} w/ condition        
        & \ding{51} & \ding{51} & \ding{55} & 2.48 & 2.55 & 2.41 & 2.55 \\
    + {\methodname} w/o condition       
        & \ding{51} & \ding{51} & \ding{51} & 22.90 & 25.15 & 24.29 & 23.68 \\
    + {\methodname} w/ condition        
        & \ding{51} & \ding{51} & \ding{51} & \textbf{2.21} & \textbf{2.12} & \textbf{2.01} & \textbf{2.08} \\
    \bottomrule
    \end{tabular}
\end{table*}

%% file: figures/afhq.tex
\begin{figure}[tbp]
\centering
\begin{tabular}{ccc}
    & VP & VE
    \\
    \rotatebox[origin=c]{90}{EDM} & 
    \begin{minipage}{0.42\linewidth}\includegraphics[width=\linewidth]{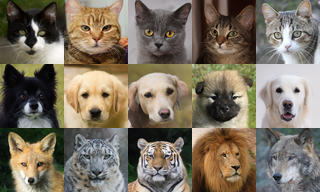}\end{minipage} & 
    \begin{minipage}{0.42\linewidth}\includegraphics[width=\linewidth]{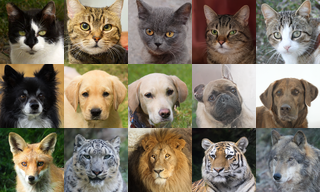}\end{minipage} 
    \\ \\

    \rotatebox[origin=c]{90}{\methodname} & 
    \begin{minipage}{0.42\linewidth}\includegraphics[width=\linewidth]{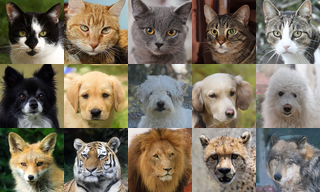}\end{minipage} & 
    \begin{minipage}{0.42\linewidth}\includegraphics[width=\linewidth]{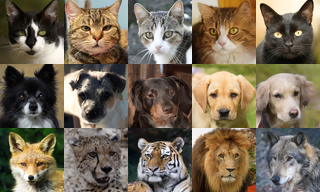}\end{minipage} 
    \\

\end{tabular}
\caption{Generated Images of our \methodname and the baseline EDM on AFHQv2.}
\label{fig:afhq}
\end{figure}

%% file: figures/overfit.tex
\begin{figure*}[t]
\centering
\subfloat[]{
\label{fig:a}
\includegraphics[width=0.33\textwidth]{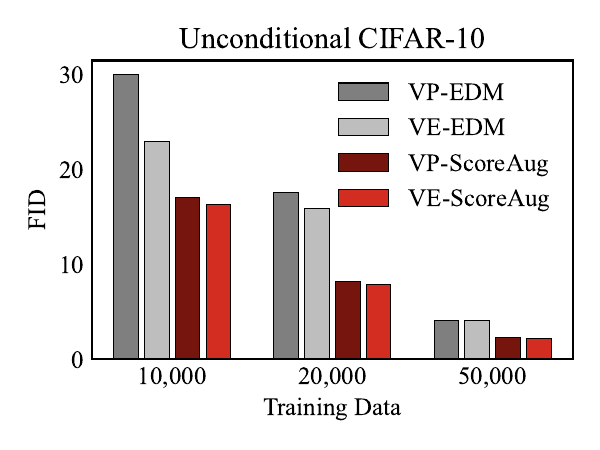}}
\subfloat[]{
\label{fig:b}
\includegraphics[width=0.33\textwidth]{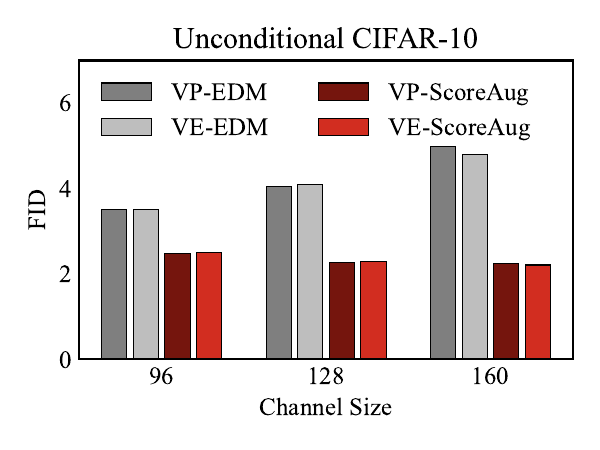}}
\subfloat[]{
\label{fig:c}
\includegraphics[width=0.33\textwidth]{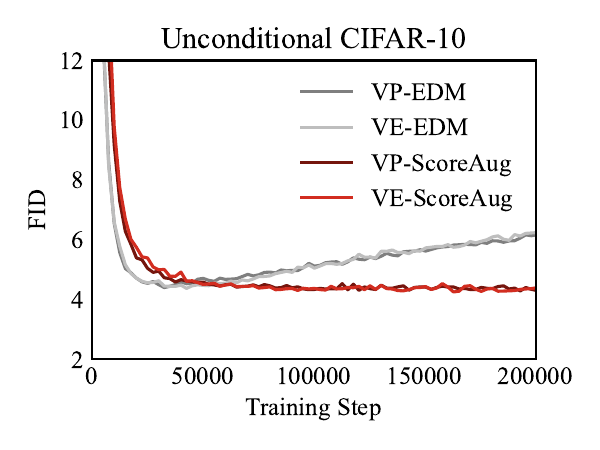}}
\\
\subfloat[]{
\label{fig:d}
\includegraphics[width=0.33\textwidth]{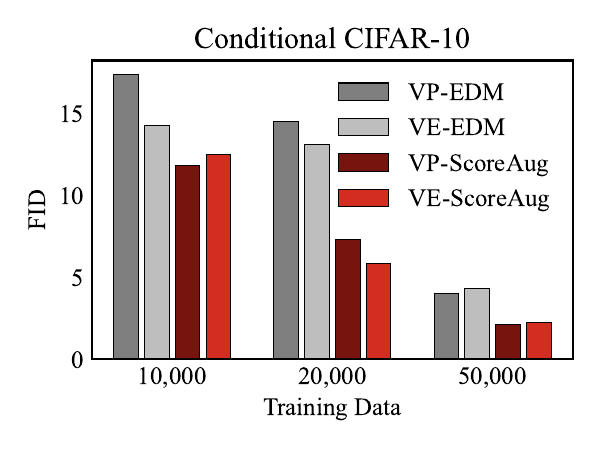}}
\subfloat[]{
\label{fig:e}
\includegraphics[width=0.33\textwidth]{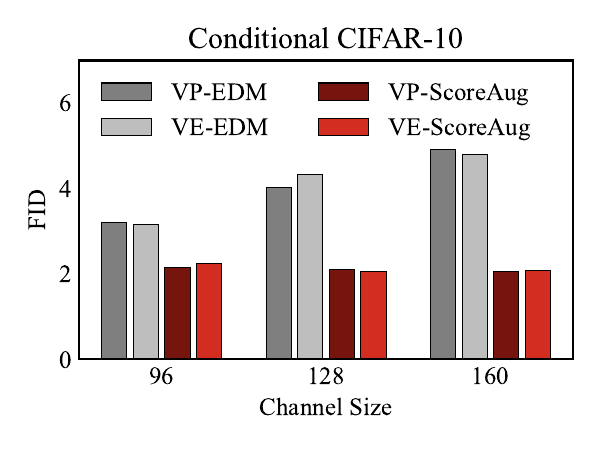}}
\subfloat[]{
\label{fig:f}
\includegraphics[width=0.33\textwidth]{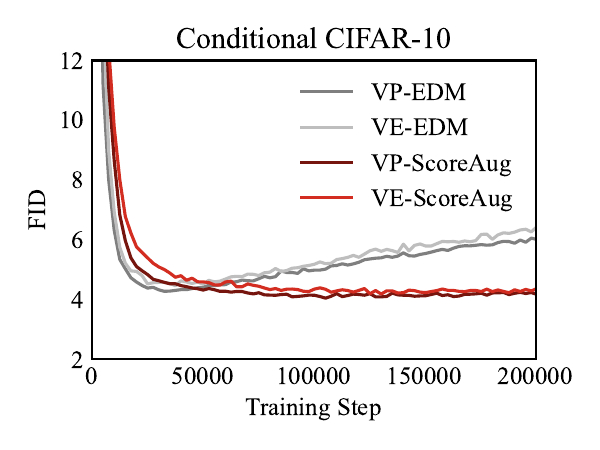}}
\caption{FID comparisons of \methodname with EDM in VP and VE diffusion formulations on unconditional and conditional CIFAR-10 datasets: (a,d) different training data sizes; (b,e) different model sizes (channels); (c,f) different training steps (FID is evaluated with 10,000 samples).}
\label{fig:6in1}
\end{figure*}

%% file: appendix.tex
\appendix

\section{Proof}
\label{app:proof}

\paragraph{Optimal Denoiser}
For the loss function in~\cref{eq:scoreaug_loss}, its expansion yields:
\begin{align*}
    &\mathcal{L}(D;\sigma,\boldsymbol{\omega}) \\
    &= \mathbb{E}_{\mathbf{d}\sim p_\mathrm{data}}\mathbb{E}_{\mathbf{n}\sim\mathcal{N}(\mathbf{0},\sigma^2\mathbf{I})}\|D(T(\mathbf{d}+\mathbf{n};\boldsymbol{\omega});\sigma,\boldsymbol{\omega}) - T(\mathbf{d};\boldsymbol{\omega})\|_2^2 \\
    &= \mathbb{E}_{\mathbf{d}\sim p_\mathrm{data}}\mathbb{E}_{\mathbf{x}\sim\mathcal{N}(\mathbf{d},\sigma^2\mathbf{I})}\|D(T(\mathbf{x};\boldsymbol{\omega});\sigma,\boldsymbol{\omega}) - T(\mathbf{d};\boldsymbol{\omega})\|_2^2 \\
    &= \mathbb{E}_{\mathbf{d}\sim p_\mathrm{data}}\mathbb{E}_{\mathbf{x}\sim\mathcal{N}(\mathbf{d},\sigma^2\mathbf{I})}\|D(\mathbf{T}\mathbf{x};\sigma,\boldsymbol{\omega}) - \mathbf{T}\mathbf{d}\|_2^2 \\
    &= \mathbb{E}_{\mathbf{y}_0\sim \hat{p}_\mathrm{data}}\mathbb{E}_{\mathbf{y}\sim\mathcal{N}(\mathbf{y}_0,\sigma^2\mathbf{T}\mathbf{T}^\top)}\|D(\mathbf{y};\sigma,\boldsymbol{\omega}) - \mathbf{y}_0\|_2^2 \\
    &= \int_{\mathbb{R}^d} \frac{1}{N}\sum_{i=1}^N \mathcal{N}(\mathbf{y};\mathbf{y}_i,\sigma^2\mathbf{T}\mathbf{T}^\top) \|D(\mathbf{y};\sigma,\boldsymbol{\omega}) - \mathbf{y}_i\|_2^2\mathrm{d}\mathbf{y}
\end{align*}
To obtain the optimal denoiser, we minimize it independently for each $\mathbf{y}$. Being a convex optimization problem, we set its derivative to zero and obtain the following.
\begin{align*}
    &0=\nabla_{D(\mathbf{y};\sigma,\boldsymbol{\omega})}\left[\mathcal{L}(D;\mathbf{y},\sigma,\boldsymbol{\omega})\right] \\
    &0=\nabla_{D(\mathbf{y};\sigma,\boldsymbol{\omega})}\left[\frac{1}{Y}\sum_{i=1}^Y\mathcal{N}(\mathbf{y};\mathbf{y}_i,\sigma^2\mathbf{T}\mathbf{T}^\top)\|D(\mathbf{y};\sigma,\boldsymbol{\omega})-\mathbf{y}_i\|_2^2\right] \\ 
    &0=\sum_{i=1}^Y\mathcal{N}(\mathbf{y};\mathbf{y}_i,\sigma^2\mathbf{T}\mathbf{T}^\top)\nabla_{D(\mathbf{y};\sigma,\boldsymbol{\omega})}\left[\|D(\mathbf{y};\sigma,\boldsymbol{\omega})-\mathbf{y}_i\|_2^2\right] \\ 
    &0= \sum_{i=1}^Y\mathcal{N}(\mathbf{y};\mathbf{y}_i,\sigma^2\mathbf{T}\mathbf{T}^\top)\left[D(\mathbf{y};\sigma,\boldsymbol{\omega})-\mathbf{y}_i\right] \\ 
    &D(\mathbf{y};\sigma,\boldsymbol{\omega})=\frac{\sum_{i=1}^Y\mathcal{N}(\mathbf{y};\mathbf{y}_i,\sigma^2\mathbf{T}\mathbf{T}^\top)\mathbf{y}_i}{\sum_{i=1}^Y\mathcal{N}(\mathbf{y};\mathbf{y}_i,\sigma^2\mathbf{T}\mathbf{T}^\top)}
\end{align*}

\paragraph{Score Function}
For the transformation $\mathbf{y}=\mathbf{T}\mathbf{x}$, its score function can be expressed as:
\begin{align}
    &\nabla_\mathbf{y}\log p(\mathbf{y};\sigma) \nonumber \\
    &= \frac{\nabla_\mathbf{y}p(\mathbf{y};\sigma)}{p(\mathbf{y};\sigma)} = \frac{\sum_{i=1}^Y\nabla_\mathbf{y}\mathcal{N}(\mathbf{y};\mathbf{y}_i,\sigma^2\mathbf{T}\mathbf{T}^\top)}{\sum_{i=1}^Y\mathcal{N}(\mathbf{y};\mathbf{y}_i,\sigma^2\mathbf{T}\mathbf{T}^\top)} \nonumber \\ 
    &= (\mathbf{T}\mathbf{T}^\top)^\dagger\left(\frac{\sum_{i=1}^Y\mathcal{N}(\mathbf{y};\mathbf{y}_i,\sigma^2\mathbf{T}\mathbf{T}^\top)\mathbf{y}_i}{\sum_{i=1}^Y\mathcal{N}(\mathbf{y};\mathbf{y}_i,\sigma^2\mathbf{T}\mathbf{T}^\top)}-\mathbf{y}\right)/\sigma^2 \label{eq:derivation} \\
    &= (\mathbf{T}\mathbf{T}^\top)^\dagger\left(D(\mathbf{y};\sigma,\boldsymbol{\omega})-\mathbf{y}\right)/\sigma^2 \nonumber
\end{align}
where~\cref{eq:derivation} comes from $\nabla_\mathbf{y}\mathcal{N}(\mathbf{y};\mathbf{y}_i,\sigma^2\mathbf{T}\mathbf{T}^\top)=\mathcal{N}(\mathbf{y};\mathbf{y}_i,\sigma^2\mathbf{T}\mathbf{T}^\top)(\mathbf{T}\mathbf{T}^\top)^\dagger(\mathbf{y}_i-\mathbf{y})/\sigma^2$.

\subsection{Transformation of Score Functions under Surjective Smooth Mappings}\label{proof:thm}
We have a random vector $\mathbf{x}\in \mathbb{R}^n$ with smooth, positive density $p(\mathbf{x};\sigma)$, and a smooth surjective map
\begin{equation*}
    T:\mathbb{R}^n \to \mathbb{R}^m, m\leq n,
\end{equation*}
of full row‑rank $m$ on the support of interest. Write $\mathbf{y}=T(\mathbf{x})$ and denote by
\begin{equation*}
    p(\mathbf{y};\sigma) = \int  p(\mathbf{x};\sigma) \delta(\mathbf{y} - T(\mathbf{x})) \mathrm{d} \mathbf{ x}.
\end{equation*}
its marginal density. 
Since everything is smooth and ppp decays at infinity, we may exchange $\nabla_{\mathbf{y}}$ and the integral:
\begin{equation*}
    \nabla_{\mathbf{y}}p(\mathbf{y};\sigma) = \int  p(\mathbf{x};\sigma) \nabla_{\mathbf{y}}\delta(\mathbf{y} - T(\mathbf{x})) \mathrm{d} \mathbf{ x}.
\end{equation*}
Next, use the chain‑rule for the delta:
\begin{equation*}
  \nabla_{\mathbf{x}} \delta(\mathbf{y}- T(\mathbf{x})) = -\mathbf{J}_{T}(\mathbf{x})^\top\nabla_{\mathbf{y}}\delta(\mathbf{y}-T(\mathbf{x})),  
\end{equation*}
where $\mathbf{J}_{T}(\mathbf{x})$ is the $m\times n$ Jacobian matrix of $T$ at $\mathbf{x}$.
Since $\mathbf{J}_{T}(\mathbf{x})$ has full row rank $m$, the $m\times m$ matrix $\mathbf{J}_T(\mathbf{x})\mathbf{J}_T(\mathbf{x})^\top$ is invertible. We can solve for $\delta(\mathbf{y}-T(\mathbf{x}))$: 
\begin{equation*}
    -(\mathbf{J}_T(\mathbf{x})\mathbf{J}_T(\mathbf{x})^\top)^{-1}\mathbf{J}_T(\mathbf{x}) \nabla_{\mathbf{x}} \delta(\mathbf{y}- T(\mathbf{x})) = \nabla_{\mathbf{y}}\delta(\mathbf{y}-T(\mathbf{x})).
\end{equation*}
Let $\mathcal{J}= (\mathbf{J}_T(\mathbf{x})\mathbf{J}_T(\mathbf{x})^\top)^{-1}\mathbf{J}_{T}(\mathbf{x})$.  Note that $\mathcal{J}$ is the transpose of Moore-Penrose pseudo-inverse $(\mathbf{J}_{T}(\mathbf{x})^{\dagger})^\top$. Substitute back into the expression for $\nabla_{\mathbf{y}} p(\mathbf{y};\sigma)$:
\begin{equation*}
   \nabla_{\mathbf{y}} p(\mathbf{y};\sigma) = -\int p(\mathbf{x};\sigma) \mathcal{J}\nabla_{\mathbf{x}} \delta(\mathbf{y}- T(\mathbf{x}))\mathrm{d}\mathbf{x} . 
\end{equation*}
Let's look at the $i$-th component of $\nabla_{\mathbf{y}} p(\mathbf{y};\sigma) $:
\begin{equation*}
   (\nabla_{\mathbf{y}}p(\mathbf{y};\sigma))_{i} = - \int p(\mathbf{x};\sigma) \sum_{k} \mathcal{J}_{ik} \frac{ \partial \delta(\mathbf{y}- T(\mathbf{x})) }{ \partial x_{k} } \mathrm{d} \mathbf{x}. 
\end{equation*}
Using integration by parts:
\begin{align*}
(\nabla_{\mathbf{y}}p(\mathbf{y};\sigma))_{i} &= - \sum_{k} \int  \frac{ \partial (p(\mathbf{x};\sigma) \mathcal{J}_{ik})}{ \partial x_{k} } \delta(\mathbf{y}- T(\mathbf{x})) \mathrm{d} \mathbf{x} \\
&= - \sum_{k} \int  \left(\frac{ \partial p(\mathbf{x};\sigma) }{ \partial x_{k} }\mathcal{J}_{ik}\right) \delta(\mathbf{y}- T(\mathbf{x})) \mathrm{d} \mathbf{x}.
\end{align*}
Let $\mathrm{Div}_\mathrm{rows} (M)_i = \sum_{k} \frac{ \partial M_{ik} }{ \partial x_{k} }$. This term represents the divergence of each row of $M$.
\begin{align*}
   &\nabla_{\mathbf{y}} \log p(\mathbf{y};\sigma) = \frac{{\nabla_{\mathbf{y}} p(\mathbf{y};\sigma)}}{p(\mathbf{y};\sigma)} \\
   &= \mathbb{E}_{p(\mathbf{x}|\mathbf{y};\sigma)}\left[\mathcal{J} \frac{\nabla_{\mathbf{y}}p(\mathbf{x};\sigma)}{p(\mathbf{x};\sigma)} + \mathrm{Div}_\mathrm{rows}(\mathcal{J}) \right] .
\end{align*}
The matrix calculus identities are known (proven later):
\begin{equation*}
    \mathrm{Div}_\mathrm{rows}(\mathcal{J}) = - \mathcal{J} \frac{1}{2} \nabla_{\mathbf{x}} \log \det(\mathbf{J}_T(\mathbf{x})\mathbf{J}_T(\mathbf{x})^\top).
\end{equation*}
Substituting this into the expression:
\begin{align*}
    &\nabla_{\mathbf{y}} \log p(\mathbf{y};\sigma) \\
    &= \mathbb{E}_{p(\mathbf{x}|\mathbf{y};\sigma)} \left[ \mathcal{J}\left( \nabla_{\mathbf{x}} \log p(\mathbf{x};\sigma) - \frac{1}{2} \nabla_{\mathbf{x}} \log \det(\mathbf{J}_T(\mathbf{x})\mathbf{J}_T(\mathbf{x})^\top) \right) \right]. 
\end{align*}
If $T$ is a global diffeomorphism then $\mathbf{J}_T(\mathbf{x})$ is square and invertible, $\mathcal{J} = \mathbf{J}_T(\mathbf{x})^{-\top}$, and
\begin{equation*}
   \frac{1}{2}\nabla_{\mathbf{x}} \log \det (\mathbf{J}_T(\mathbf{x})\mathbf{J}_T(\mathbf{x})^\top) = \nabla_{\mathbf{x}} \log |\det \mathbf{J}_T(\mathbf{x})|. 
\end{equation*}
Hence,
\begin{align*}
    &\nabla_{\mathbf{y}} \log p (\mathbf{y};\sigma) \\
    &= \mathbf{J}_T(\mathbf{x})^{-\top}\left( \nabla_{\mathbf{x}} \log p(\mathbf{x};\sigma) - \nabla_{\mathbf{x}} \log |\det(\mathbf{J}_T(\mathbf{x}))| \right).
\end{align*}
If $T(\mathbf{x})=\mathbf{Tx}$, where $\mathbf{T}$ is a constant $m\times n$ matrix with full row rank $m$:
\begin{equation*}
    \mathcal{J} = (\mathbf{T}\mathbf{T}^\top)^{-1}\mathbf{T}, \frac{1}{2}\nabla_{\mathbf{x}} \log \det (\mathbf{J}_T(\mathbf{x})\mathbf{J}_T(\mathbf{x})^\top) = 0.
\end{equation*}
Hence,
\begin{equation*}
    \nabla_{\mathbf{y}} \log p(\mathbf{y};\sigma) = (\mathbf{T}\mathbf{T}^\top)^{-1}\mathbf{T} \cdot \mathbb{E}_{p(\mathbf{x}|\mathbf{y};\sigma)} \left[\nabla_{\mathbf{x}} \log p(\mathbf{x};\sigma)\right].
\end{equation*}

\paragraph{The matrix calculus identities}
Since the Jacobian determinant is defined on a smooth function, satisfying $\partial_{i} \mathbf{J}_{kj} = \partial_{j} \mathbf{J}_{ki}$, we define the right inverse $\mathbf{J}^{\dagger} = \mathbf{J}^\top (\mathbf{J}\mathbf{J}^\top)^{-1}$. Prove that:
\begin{equation*}
    \mathrm{Div}_\mathrm{rows}((\mathbf{J}^{\dagger})^\top) = - \mathbf{J}^{\dagger \top} \frac{1}{2} \nabla_{\mathbf{x}} \log \det(\mathbf{J}\mathbf{J}^\top).
\end{equation*}
This is equivalent to proving:
\begin{equation*}
    \mathbf{J}^\top \mathrm{Div}_\mathrm{rows}((\mathbf{J}^{\dagger})^\top) = - \frac{1}{2} \nabla_{\mathbf{x}} \log \det (\mathbf{J}\mathbf{J}^{\top}).
\end{equation*}
Consider the $k$-th term on the left:
\begin{equation*}
    \text{LHS}_{k} = \sum_{i} \mathbf{J}_{ik} \left( \sum_{j} \frac{ \partial ((\mathbf{J}^{\dagger})^\top)_{ij} }{ \partial x_{j} } \right) = \sum_{ij} \mathbf{J}_{ik} \partial_{j} \mathbf{J}^{\dagger}_{ji}.
\end{equation*}
The $k$-th term on the right side:
\begin{align*}
\text{RHS}_{k} &= -\frac{1}{2}\frac{ \partial \ln \det (\mathbf{J}\mathbf{J}^{\top}) }{ \partial x_{k} } \\
&= -\frac{1}{2}\sum_{ij} \left(\frac{ \partial \ln \det (\mathbf{J}\mathbf{J}^{\top}) }{ \partial \mathbf{J} } \right)_{ij} \frac{ \partial \mathbf{J}_{ij} }{ \partial x_{k} } \\
&= -\sum_{ij} (\mathbf{J}^{\dagger\top})_{ij} \frac{ \partial \mathbf{J}_{ij} }{ \partial x_{k} } = -\sum_{ij} \mathbf{J}^{\dagger}_{ji} \partial_{j} \mathbf{J}_{ik}.
\end{align*}
Hence,
\begin{align*}
\text{LHS}_{k}- \text{RHS}_{k} &= \sum_{ij} \mathbf{J}_{ik} \partial_{j} \mathbf{J}^{\dagger}_{ji} + \mathbf{J}^{\dagger}_{ji} \partial_{j} \mathbf{J}_{ik} \\
&= \sum_{ij} \partial_{j} (\mathbf{J}_{ji}^{\dagger} \mathbf{J}_{ik}) \\
&= \sum_{j} \partial_{j} \left( \sum_{i} \mathbf{J}_{ji}^{\dagger} \mathbf{J}_{ik} \right) = 0,
\end{align*}
where $\mathbf{J}^{\dagger}\mathbf{J}$ is the projection operator, which exhibits zero divergence, so the above formula holds.

\section{Additional Experimental Settings}
\label{sec:exp_augs}

\paragraph{Experimental Resources}
Our all experiments were performed on a cluster of 8 NVIDIA V100 or 8 H800 GPUs. Training time and resources do not increase significantly compared to the base model.

\paragraph{Code}
Our code will be open-source upon acceptance.

\paragraph{Used Augmentations} 
Below are the data augmentation methods (with hyperparameters in parentheses) used by \methodname in~\cref{tab:main_results,tab:dit}. During each training session, one augmentation method is randomly selected with equal probability across all types, where "identity" denotes unchanging the input samples. Notably, our approach operates effectively on both pixel and latent spaces, making it fully compatible with SiT models.
\begin{itemize}
    \item \textbf{\methodname on EDM w/o NLA}
    \item \begin{itemize}
        \item \underline{Unconditional CIFAR-10}: brightness ($B=2$), translation ($R_t=0.25$), cutout ($R_c=0.5$), rotation.
        \item \underline{Conditional CIFAR-10}: brightness ($B=2$), translation ($R_t=0.25$), cutout ($R_c=0.5$), rotation.
        \item \underline{FFHQ}: translation ($R_t=0.25$), cutout ($R_c=0.5$), rotation.
        \item \underline{AFHQ}: translation ($R_t=0.25$), cutout ($R_c=0.5$), rotation.
    \end{itemize}
    \item \textbf{\methodname on EDM w/ NLA}
    \item \begin{itemize}
        \item \underline{Unconditional CIFAR-10}: translation ($R_t=0.125$), cutout ($R_c=0.25$).
        \item \underline{Conditional CIFAR-10}: translation ($R_t=0.125$), cutout ($R_c=0.25$).
        \item \underline{FFHQ}: identity, translation ($R_t=0.125$), cutout ($R_c=0.25$).
        \item \underline{AFHQ}: brightness ($B=2$), translation ($R_t=0.125$), cutout ($R_c=0.25$).
    \end{itemize}
    \item \textbf{\methodname on SiT}, \underline{ImageNet}: translation ($R_t=0.0325$)
\end{itemize}

\section{More Qualitative Results}
\label{app:more_results}
Below are visualization results on CIFAR-10 (rotation-controllable results) and FFHQ.

\input{figures/rotation_uncond_cifar10}

\input{figures/rotation_cond_cifar10}

\input{figures/ffhq}

%% file: figures/rotation_uncond_cifar10.tex
\begin{figure*}[t]
\centering
\subfloat[VP-ScoreAug, rotation=$0^\circ$]{
\includegraphics[width=0.4\linewidth]{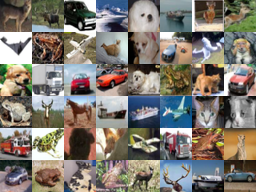}}
\subfloat[VE-ScoreAug, rotation=$0^\circ$]{
\includegraphics[width=0.4\linewidth]{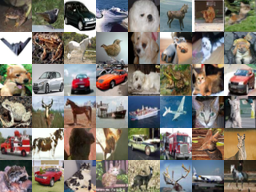}}
\\
\subfloat[VP-ScoreAug, rotation=$90^\circ$]{
\includegraphics[width=0.4\linewidth]{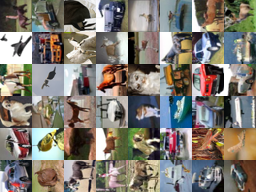}}
\subfloat[VE-ScoreAug, rotation=$90^\circ$]{
\includegraphics[width=0.4\linewidth]{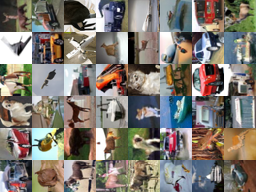}}
\\
\subfloat[VP-ScoreAug, rotation=$180^\circ$]{
\includegraphics[width=0.4\linewidth]{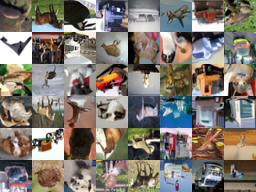}}
\subfloat[VE-ScoreAug, rotation=$180^\circ$]{
\includegraphics[width=0.4\linewidth]{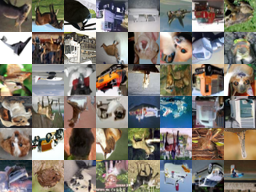}}
\\
\subfloat[VP-ScoreAug, rotation=$270^\circ$]{
\includegraphics[width=0.4\linewidth]{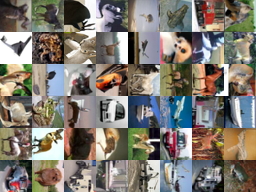}}
\subfloat[VE-ScoreAug, rotation=$270^\circ$]{
\includegraphics[width=0.4\linewidth]{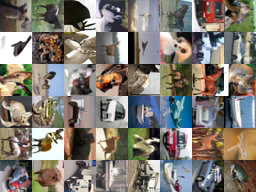}}
\caption{Augmentation-conditional generated images on unconditional CIFAR-10 of \methodname.}
\label{fig:rot_uncond_cifar10}
\end{figure*}

%% file: figures/rotation_cond_cifar10.tex
\begin{figure*}[t]
\centering
\subfloat[VP-ScoreAug, rotation=$0^\circ$]{
\includegraphics[width=0.4\linewidth]{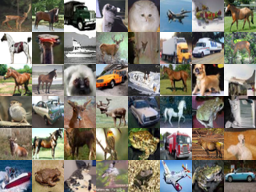}}
\subfloat[VE-ScoreAug, rotation=$0^\circ$]{
\includegraphics[width=0.4\linewidth]{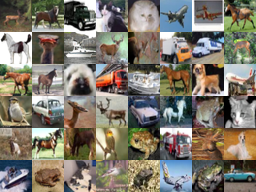}}
\\
\subfloat[VP-ScoreAug, rotation=$90^\circ$]{
\includegraphics[width=0.4\linewidth]{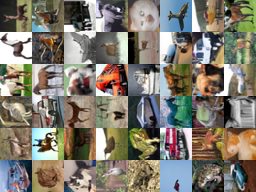}}
\subfloat[VE-ScoreAug, rotation=$90^\circ$]{
\includegraphics[width=0.4\linewidth]{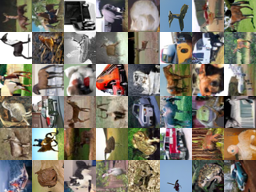}}
\\
\subfloat[VP-ScoreAug, rotation=$180^\circ$]{
\includegraphics[width=0.4\linewidth]{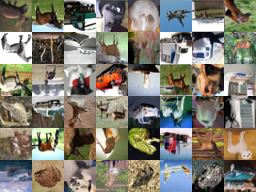}}
\subfloat[VE-ScoreAug, rotation=$180^\circ$]{
\includegraphics[width=0.4\linewidth]{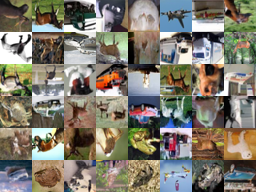}}
\\
\subfloat[VP-ScoreAug, rotation=$270^\circ$]{
\includegraphics[width=0.4\linewidth]{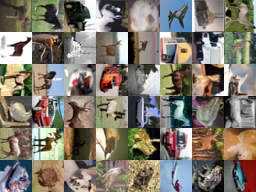}}
\subfloat[VE-ScoreAug, rotation=$270^\circ$]{
\includegraphics[width=0.4\linewidth]{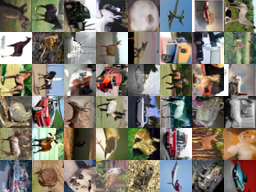}}
\caption{Augmentation-conditional generated images on conditional CIFAR-10 of \methodname.}
\label{fig:rot_cond_cifar10}
\end{figure*}

%% file: figures/ffhq.tex
\begin{figure*}[t]
\centering
\subfloat[VP-EDM w/o NLA]{
\includegraphics[width=0.4\linewidth]{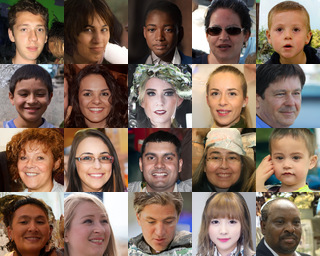}}
\subfloat[VE-EDM w/o NLA]{
\includegraphics[width=0.4\linewidth]{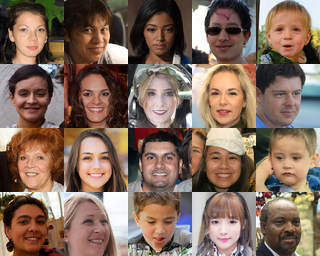}}
\\
\subfloat[VP-ScoreAug w/o NLA]{
\includegraphics[width=0.4\linewidth]{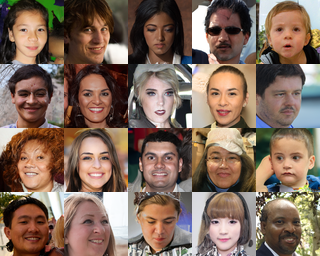}}
\subfloat[VE-ScoreAug w/o NLA]{
\includegraphics[width=0.4\linewidth]{assets/ffhq/ffhq_ddpmpp_scoreaug.png}}
\\
\subfloat[VP-EDM w/ NLA]{
\includegraphics[width=0.4\linewidth]{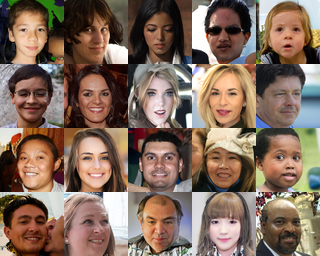}}
\subfloat[VE-EDM w/ NLA]{
\includegraphics[width=0.4\linewidth]{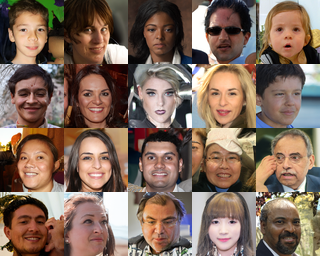}}
\\
\subfloat[VP-ScoreAug w/ NLA]{
\includegraphics[width=0.4\linewidth]{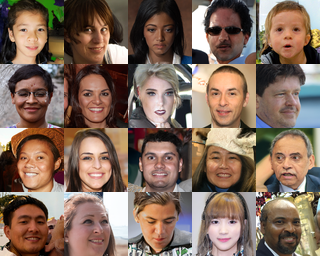}}
\subfloat[VE-ScoreAug w/ NLA]{
\includegraphics[width=0.4\linewidth]{assets/ffhq/ffhq_ddpmpp_nla_scoreaug.png}}
\label{fig:ffhq}
\caption{Generated images of EDM and \methodname without and with NLA on FFHQ.}
\end{figure*}